\documentclass[manuscript]{geophysics}
\usepackage{hyperref}
\usepackage{multirow}
\usepackage{array}
\usepackage{amsmath}
\usepackage{amssymb}
\usepackage{lineno}
\usepackage[table]{xcolor}

\newcolumntype{L}[1]{>{\raggedright\arraybackslash}p{#1}}

\begin{document}

\begin{titlepage}
    \centering
    
    {\Large Domain-Guided Prompting of the Segment Anything Model for Seismic Interpretation: The Role of Attributes, Visualization, and Hybrid Prompts\par}

    \vspace{1cm}
    
    {\large Aniq Ahmad$^{1}$, Heather Bedle$^{1}$, Ahmad Mustafa$^{2}$\par}
    
    \vspace{0.8cm}
    
    {\normalsize $^{1}$School of Geosciences, University of Oklahoma, Norman, Oklahoma, USA\par}
    {\normalsize $^{2}$King Fahd University of Petroleum and Minerals, Dhahran, Saudi Arabia\par}
    {\small
    \textit{This manuscript has been submitted to Geophysics for possible publication and is currently under peer review.}
    \par}
    \vfill
\end{titlepage}

\begin{abstract}

The advent of large pretrained foundation models for computer vision has significantly improved the efficiency of visual data interpretation. The Segment Anything Model (SAM), in particular, offers powerful zero-shot segmentation capabilities through prompt-based interaction, thus making it a promising tool for seismic interpretation. However, most existing applications of SAM rely on fine-tuning for specific geological targets, which requires extensive labeled data, incurs high computational cost, and often compromises the model’s generalization capability. In this study, we introduce a principled framework for zero-shot adaptation of foundation models to seismic data. The framework is built on two key components: (1) aligning seismic attributes and visualization choices (e.g., colormaps) with the geological target of interest, and (2) employing a hybrid prompting strategy that combines sparse user-defined point prompts with dense mask prompts derived from SAM’s internal feature activations. We systematically evaluate this framework across multiple geological targets, datasets, prompt configurations, and seismic attribute representations. Our results demonstrate that geologic-target-aware selection of seismic attributes and colormaps, combined with hybrid prompting, enhances the separability of geological features and improves boundary delineation and segmentation accuracy relative to point-based prompting alone. Our findings show that, when these components are jointly applied, SAM can achieve competitive segmentation performance in a fully zero-shot setting, thereby eliminating the need to retrain SAM for each geologic feature. This work establishes a practical and scalable pathway to leverage foundation models in seismic interpretation, reducing reliance on labeled data while preserving model generality.

\end{abstract}

\section{Introduction}
Seismic interpretation is a critical component of hydrocarbon exploration, geological hazard assessment, and subsurface resource management. Traditionally, this process relies on manual interpretation by geoscientists, rendering it both time-intensive and inherently subjective, particularly for large-scale seismic datasets. These limitations motivated the adoption of automated approaches to improve efficiency, consistency, and scalability.

Recent advances in artificial intelligence, particularly in computer vision, have enabled significant progress toward semi-automated and fully automated seismic interpretation. However, most supervised deep learning approaches require large volumes of labeled training data to achieve robust performance. In practice, geological targets vary widely, from high-contrast structures such as salt bodies to subtle stratigraphic features like channels, often requiring task-specific datasets and independently trained models. This reliance on curated annotations introduces a substantial bottleneck, as labeling seismic data is both costly and labor-intensive, ultimately limiting model scalability and generalization. Despite these challenges, prior work has demonstrated success in specific tasks, including salt body classification \citep{waldeland2018convolutional,lubo2021exhaustive}, fault detection \citep{wu2018convolutional,mustafa_visual_fault}, seismic imaging \citep{ma2024deep}, channel identification\citep{pham2019automatic}, and facies classification \citep{liu2020seismic,mustafa2023active,tahir2024comparing}. Nevertheless, the effectiveness of these models remains tightly coupled to the availability of large task-specific labeled data.

Beyond data limitations, conventional deep learning models exhibit two fundamental shortcomings. First, they are typically designed to predict a fixed set of predefined classes, restricting their applicability to narrowly defined tasks. Secondly, the geological variability across basins leads to diverse and often ambiguous feature representations in seismic interpretation, thereby limiting model generalization and reuse. Furthermore, the lack of interactivity in these models prevents the incorporation of user feedback during inference. Consequently, geoscientists cannot iteratively refine model outputs using domain expertise, an essential component of expert-driven interpretation workflows.

Foundation models have recently emerged as a promising paradigm to address these challenges. Often built upon transformer-based architectures \citep{dosovitskiy2020image,liu2021swin}, these models leverage large-scale self-supervised pretraining on unlabeled image corpora to learn generalizable visual representations. Subsequent fine-tuning enables adaptation to downstream tasks with reduced labeled data requirements. However, early applications of foundation models to seismic interpretation \citep{saha2025adasemseg, pham2025seisbert} continue to rely on domain-specific labeled datasets, thereby inheriting many of the same data dependency limitations.

The introduction of the family of Segment Anything Models (SAM 1, SAM 2, and SAM 3) \citep{Kirillov_2023_ICCV,ravi2024sam,carion2025sam} by Meta represents a significant advancement in interactive image segmentation. Trained on a massive corpus of images and segmentation masks, both these models achieve strong zero-shot performance across diverse natural image domains. Their promptable design enables users to guide segmentation at inference time through inputs such as points, bounding boxes, or masks, thereby integrating human expertise directly into the segmentation process without requiring task-specific retraining.

In the context of seismic interpretation, this paradigm offers a compelling solution to the data bottleneck. SAM allows interpreters to provide prompts on selected sections (e.g., inlines, cross lines, or timeslices), which can then be used to generate dense segmentations across large volumetric datasets. This interaction model significantly amplifies the impact of limited user input, reducing annotation effort while enabling scalable interpretation across multiple geological features and datasets.

However, the direct application of SAM to seismic data remains challenging due to fundamental differences between natural images and seismic imagery. Unlike natural images, seismic data lack well-defined object boundaries and exhibit complex, texture-dominated patterns governed by subsurface physics. Geological features often manifest as subtle variations in amplitude, continuity, or phase rather than distinct objects with clear edges. Furthermore, seismic images are typically single-channel representations with domain-specific visual characteristics, whereas SAM is trained on multi-channel natural images with rich color and semantic cues. These discrepancies limit the effectiveness of standard prompting strategies when applied directly to seismic data.

Consequently, existing approaches that apply SAM for seismic feature segmentation \citep{chen2024seismic, atolagbe2025towards} often rely on computationally expensive retraining and fine-tuning \citep{ahmad2026interactive} using labeled seismic datasets from the target domain. While this improves task-specific performance, it transforms SAM into a highly specialized model, diverging from its original design as a general-purpose segmentation framework. This approach introduces several key challenges. First, it reintroduces data dependency, as large volumes of manually annotated seismic labels are required, which are costly and subject to interpreter bias. Second, fine-tuning large-scale foundation models is computationally intensive, requiring substantial resources and limiting practical scalability across multiple use cases. Third, task-specific finetuning reduces the model’s ability to generalize to unseen features, as it becomes overly specialized to the training distribution. This limitation can be attributed to catastrophic forgetting, a phenomenon in which updating a neural network with new data compromises previously learned representations \citep{kirkpatrick2017overcoming}.


In this work, we propose a systematic zero-shot adaptation framework for seismic segmentation using SAM, demonstrating that competitive performance can be achieved by integrating geologic target-aware seismic attribute representations and hybrid prompting, without retraining. Specifically, we explore three complementary directions.

First, we examine the limitations of conventional point and box-based prompts for seismic feature segmentation. Due to the absence of clearly defined object boundaries, sparse prompts often under-specify the target region. To address this, we propose a two-step prompting strategy in which the model’s initial segmentation logits are fed back as a dense mask prompt, together with the original sparse inputs. This strategy leverages SAM’s internal feature representations to refine segmentation outputs and forms a key component of our zero-shot adaptation framework for improving accuracy under sparse supervision.

Second, we explore the use of seismic attributes to enhance feature visibility and improve prompt effectiveness. We incorporate attribute-enhanced representations, including multi-spectral coherence,  relative acoustic impedance, and instantaneous frequency, which emphasize structural discontinuities, stratigraphic changes, and frequency-based textural variations, respectively. These attributes improve the visual separability of geological structures such as channels, facies, and salt boundaries, thereby providing more informative inputs for prompt-based segmentation. This component leverages domain-specific feature enhancement to better align seismic data with the visual patterns that foundation models are designed to interpret, making seismic attribute-target matching a second key element of our framework.

Finally, we explore the impact of color rendering strategies. Although seismic data are inherently single-channel, they can be visualized using different colormaps, which influence the perceptual prominence of geological features. We evaluate multiple colormap configurations to assess their effect on SAM’s segmentation performance. Together with attribute selection and hybrid prompting, these visualization choices help define a systematic zero-shot workflow for adapting SAM to seismic data.

Through a comprehensive evaluation across diverse datasets, geological targets, prompt types, seismic attributes, and visualization strategies, this study establishes a systematic framework for the zero-shot application of SAM to seismic interpretation. By avoiding geologic feature-specific fine-tuning and instead leveraging domain-informed input transformations, seismic attributes, colormaps aligned with geologic targets, and hybrid prompting, we demonstrate a practical pathway for deploying foundation models in real-world geophysical workflows without retraining. Ultimately, we position SAM as an interactive assistant that complements geoscientists' expertise, enabling efficient, scalable, and interpretable segmentation of geological features on seismic data.

\section*{Theory}

\subsection{Model Architecture}

The SAM architecture comprises an image encoder, a prompt encoder, and a mask decoder. Conceptually, the SAM architecture, as shown in workflow Figure \ref{fig:fig_1}, is a two-stage process: the model first converts the input image into a compact feature representation and subsequently uses user guidance or prompts to generate a segmentation mask from those features.
The imagee encoder is a Vision Transformer (ViT). Conceptually, the ViT partitions the input image into fixed-size patches, often known as tokens, and then learns relationships among these tokens to build a global representation of the scene. The encoder produces a latent feature map at approximately one-sixteenth the input resolution, reducing computational cost while retaining key structural patterns. The ViT backbone is pre-trained on a very large collection of natural images using a masked autoencoding (MAE) objective, where the model learns to reconstruct missing patches from visible ones. This pretraining encourages learning general visual cues such as edges, textures, and shapes, which can transfer to other image types. In the seismic setting, these cues correspond to amplitude textures, discontinuities, and structural patterns that are useful for delineating geological features.

The prompt encoder transforms user guidance into numerical representations, namely embeddings, that lie in the same latent space as the image features. SAM supports multiple prompt types, grouped into sparse and dense prompt categories. Sparse prompts include point clicks, defined by $(x, y)$ coordinates and a foreground/background label, and bounding boxes that constrain the region of interest. Dense prompts include mask prompts, which are low-resolution logits typically obtained from a first or previous iteration. They provide spatial guidance over a larger area.

Mask decoder is a lightweight Transformer that fuses the image feature map and the prompt embeddings to predict segmentation masks. It uses cross-attention, which can be understood as a mechanism that allows the prompt information to query the image feature map and focus on regions most consistent with the provided guidance. Importantly, SAM typically produces multiple candidate masks for a given prompt, along with a predicted quality score for each mask, which is often interpreted as a confidence or quality estimate. We select the highest-scoring mask as the final segmentation.

In our study, we utilize SAM in an image-based setting, where each seismic inline, crossline, or time slice is treated as a single input image and segmented independently.

\subsection{Two-Step Prompting Mechanism}

We define a set of sparse user prompts as $\mathcal{P} = \{(x_i, y_i, l_i)\}_{i=1}^{N}$ where $(x_i, y_i)$ denotes the spatial location of the $i$-th prompt and $l_i \in \{+1, 0\}$
indicates positive (foreground) or negative (background) labels. The prompt encoder maps these inputs to an embedding:
\begin{equation}
E_p = \Phi_p(\mathcal{P}),
\label{eq:prompt_enc}
\end{equation}

where $\Phi_p(\cdot)$ denotes the prompt encoding function.  

The input seismic image $I \in \mathbb{R}^{H \times W \times C}$ is passed through the image encoder:

\begin{equation}
E_I = \Phi_I(I),
\label{eq:img_enc}
\end{equation}

where $E_I \in \mathbb{R}^{h \times w \times d}$ represents the latent image embedding.

The mask decoder predicts an initial segmentation mask:
\begin{equation}
M^{(1)} = \Phi_d(E_I, E_p),
\label{eq:postsigmoid_mask}
\end{equation}

where $M^{(1)} \in [0,1]^{H \times W}$ is obtained after upsampling and sigmoid activation.

We also extract the corresponding pre-sigmoid logits:
\begin{equation}
Z^{(1)} \in \mathbb{R}^{H \times W}.
\label{eq:logits}
\end{equation}

The dense mask prompt is combined with the original sparse prompts and fed back into the decoder:
\begin{equation}
M^{(2)} = \Phi_d(E_I, E_p, Z^{(1)}),
\label{eq:refined}
\end{equation}
where $M^{(2)} \in [0,1]^{H \times W}$ is the refined segmentation mask.

\subsection{Attribute Enhancement}

Seismic attributes are quantitative measurements derived from seismic data that help highlight specific geological features or rock properties that may not be evident in the raw amplitude data. They are designed to highlight structural and stratigraphic patterns. For example, the coherence attribute highlights discontinuities and edges, thereby improving the delineation of faults, channels, submarine canyons, karst collapses, and mass-transport complexes \citep{chopra2018coherence}. Instantaneous frequency is a single-trace attribute sensitive to changes in the spectral content of the seismic signal that reveals contrasting character inside and outside salt bodies, making it useful for salt boundary interpretation and segmentation \citep{halpert2008salt}. For facies interpretation, we use relative acoustic impedance to emphasize stratigraphic layering and impedance contrasts, as this attribute is effective in highlighting laterally and vertically varying stratigraphic units and thin-bed features \citep{hart2008stratigraphically, chopra2009relative}.

We computed seismic attributes using the Attribute Assisted Seismic Processing and Interpretation (AASPI) software. To generate coherence, we first computed dip using the gradient structure tensor method \citep{luo2006computation}, and then calculated coherence using the energy-ratio similarity algorithm \citep{gersztenkorn1999eigenstructure}. Multi-spectral coherence was computed by applying coherence across three spectral bands (10–30 Hz, 30–50 Hz, and 50–70 Hz). This increases sensitivity to subtle frequency-dependent features and yields enhanced imaging of channel incisions while reducing incoherent noise compared with single-band approaches \citep{li2018multispectral}. For salt-body interpretation, we calculated the instantaneous frequency attribute following the algorithm given by \citet{taner1979complex}. Lastly, for facies interpretation, the relative acoustic impedance attribute was computed in AASPI using the \texttt{relative\_acoustic\_impedance} program. Each attribute was treated as a single-channel image, normalized to $[0, 1]$ for SAM compatibility.

\subsection{Colormap Transformation}

We also explored color mapping as a method to enhance subtle amplitude variations. Standard interpretation colormaps (e.g., \textit{RdBu}, \textit{seismic}) were applied to raw amplitude volumes to generate high-contrast visual representations. The colormap transformations operate by redistributing amplitude values across a perceptual color gradient; they are expected to benefit targets whose boundaries are physically expressed as strong amplitude contrasts while potentially degrading performance for stratigraphic targets whose defining signal is lateral spatial continuity rather than amplitude magnitude. These transformed color-mapped images were converted to RGB, linearly rescaled to $[0, 1]$ and replicated three times to produce a three-channel image before SAM processing.

\subsection{Evaluation}

We quantitatively assessed model performance under efficient prompting and across different input configurations using precision, recall, and the F1-score, calculated from the numbers of true positives ($TP$), false positives ($FP$), and false negatives ($FN$). Precision quantifies how many predicted positive samples are correct, recall quantifies how many actual positive samples are recovered, and the F1-score summarizes both using the harmonic mean.
\citep{sokolova2009systematic}:
\begin{equation}
\mathrm{Precision}=\frac{TP}{TP+FP}
\end{equation}
\begin{equation}
\mathrm{Recall}=\frac{TP}{TP+FN}
\end{equation}
\begin{equation}
\mathrm{F1}=2\cdot\frac{\mathrm{Precision}\cdot\mathrm{Recall}}{\mathrm{Precision}+\mathrm{Recall}}
\end{equation}

In the section below, we evaluate the performance of SAM  across the different datasets. We first characterize the datasets used for benchmarking and then systematically evaluate how different prompting strategies and input configurations, ranging from raw amplitude to seismic attribute-guided enhancements, color mapping, and prompt density, influence segmentation accuracy.

\begin{figure}
    \centering
    \includegraphics[width=1\linewidth]{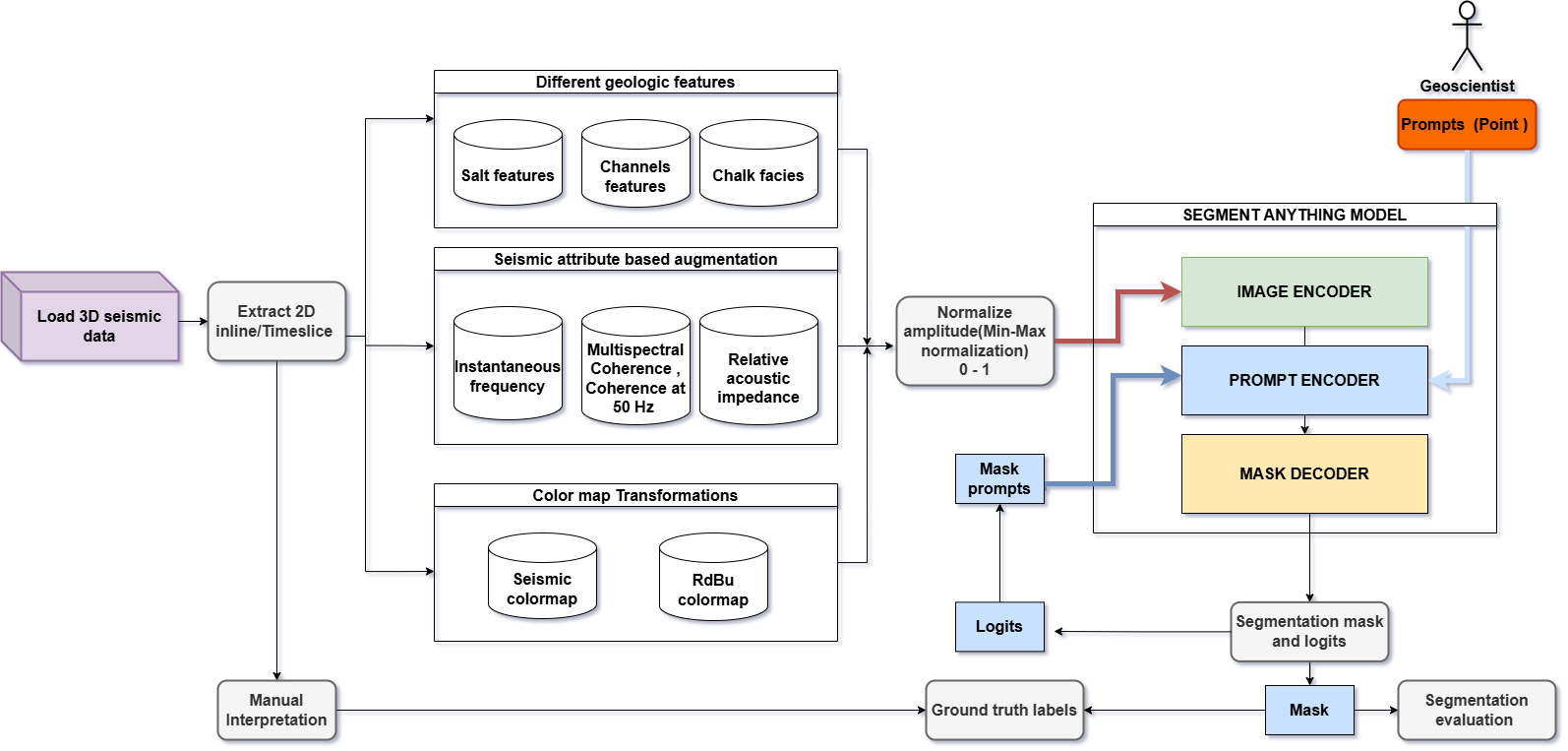}
    \caption{Overview of the SAM-based seismic segmentation workflow. Seismic slices, attribute-enhanced inputs and colormap transformations are normalized and combined with user-defined point prompts and mask prompts derived from logits (Hybrid prompting). Evaluation of model performance is carried out by taking the highest score segmentation mask from SAM and the ground truth interpretation generated by an expert geophysicist.}
    \label{fig:fig_1}
\end{figure}

The detailed workflow is illustrated in Figure~\ref{fig:fig_1}. Quantitative results in terms of precision, recall, and F1 score are reported in Table \ref{tab:prompt-attribute-color}.
\section{Results}

We evaluated the proposed workflow on three seismic datasets representing diverse geological environments. These volumes were selected to assess the model’s performance across three distinct feature classes: high and low-contrast structural bodies (salt), complex stratigraphic systems (channels), and regional facies boundaries.

\begin{figure}
    \centering
    \includegraphics[width=0.75\linewidth]{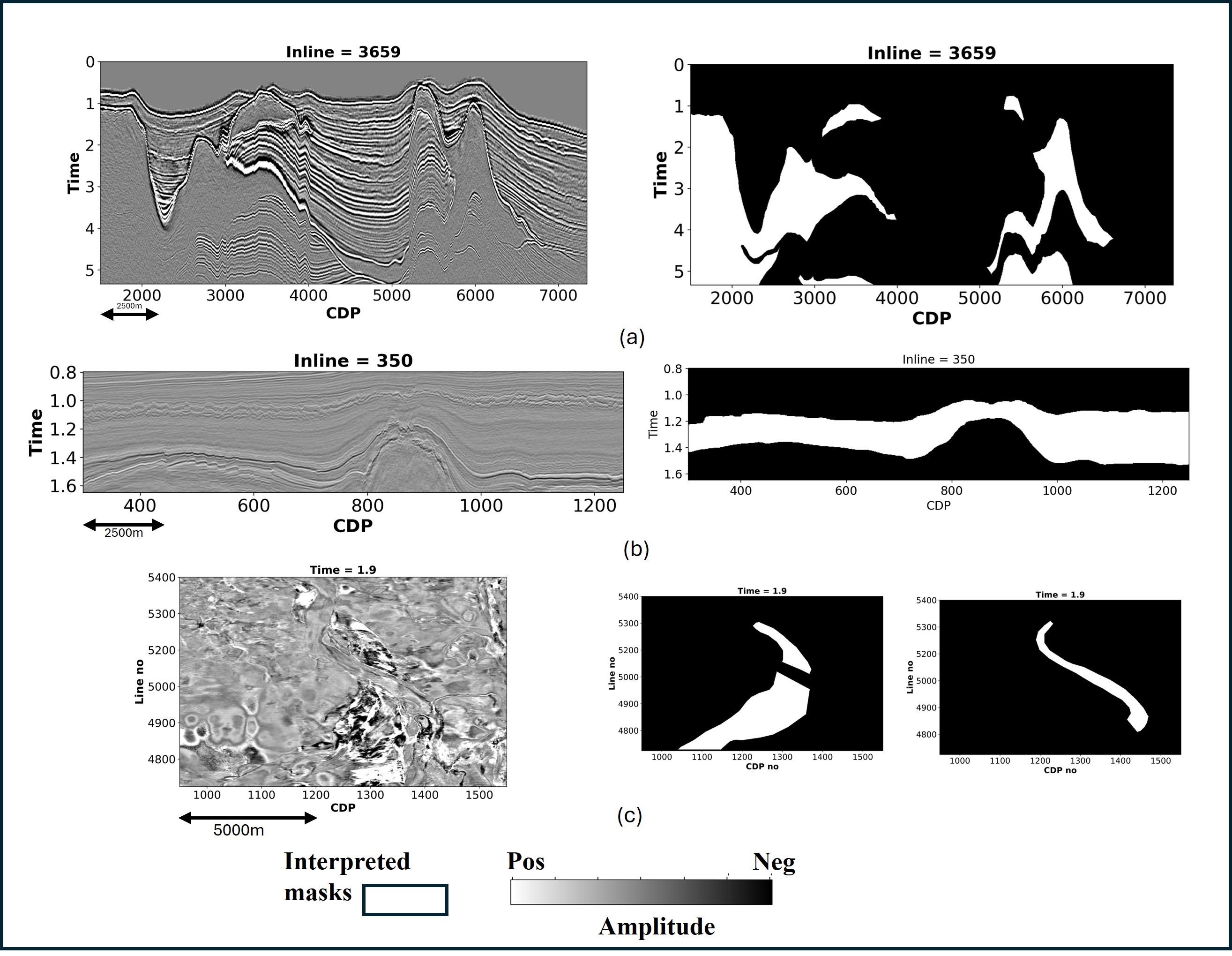}
    \caption{Interpretations generated manually for the three datasets used. Polygons are traced for dataset (a) Seam AI Interpretation phase 1 for salt bodies, (b)  F-3 Netherlands dataset for Lower North Sea Group facies, (c) Waka-3D seismic dataset for channels (mud-filled and sand-filled).}
    \label{fig:fig_2}
\end{figure}

The first dataset (Figure \ref{fig:fig_2}a) is the SEAM Phase I Interpretation Challenge volume. This synthetic dataset was generated using rigorous numerical modeling to simulate the deep water environment of the Gulf of Mexico, characterized by complex salt canopies and fine-scale stratigraphy \citep{SEAM_Interp_Challenge_2014}. We used this volume primarily for salt body segmentation.

The second dataset (Figure \ref{fig:fig_2}b) is the F-3 Block located on the Dutch North Sea continental shelf. This 16 × 24 km² area is structurally situated on the boundary between the Step Graben and the Dutch Central Graben. The region comprises ten recognized lithostratigraphic units ranging from Miocene-Quaternary claystones and sandstones to Upper Carboniferous siliceous basement rocks \citep{scheck2005crustal, duin2006subsurface}. The local geology is heavily influenced by repeated salt diapirism and Jurassic-scale subsidence \citep{remmelts1996salt}. We used this volume to segment Lower North Sea Group facies.

The third dataset (Figure \ref{fig:fig_2}c) originates from the Waka-3D survey in the Canterbury Basin, offshore New Zealand. Situated at the transition between the continental slope and rise, this area is defined by a Miocene turbidite system featuring abundant paleocanyons and deep water amplitude varying mud and sand filled channel complexes \citep{zhao2016characterizing}. These features were deposited during a tectonically driven transgressive-regressive cycle \citep{uruski2010new}. The mapped features in our study resemble amalgamated channel systems, where individual channel elements reflect shifting depositional settings over time \citep{gibson1994field}. We used this volume to segment high-amplitude sand-filled and low-amplitude mud-filled channel bodies.

To quantitatively evaluate model performance, we established ground-truth labels through manual interpretation. Facies and salt bodies were delineated by digitizing polygons on inline sections. For channel features, we picked a reference horizon and flattened the seismic volume to extract stratigraphic time slices, on which two distinct channels were mapped. An expert geophysicist validated all interpretations to ensure geological accuracy. We then converted these manual picks into binary masks to serve as ground truth, as shown in Figure \ref{fig:fig_2}(a, b and c). 

In the first set of experiments, we evaluated the impact of different prompt types on segmentation accuracy for three target classes: salt bodies, low-amplitude mud-filled channels, and Lower North Sea Group facies. In addition to the traditional bounding box and point-based prompts, we also tested our improvised strategy of using the model's internal logits as a dense mask prompt input to itself, along with the original point prompts. For these experiments, we constructed RGB input images by duplicating each original grayscale seismic slice into three image channels.

\begin{figure}
    \centering
    \includegraphics[width=0.7\linewidth]{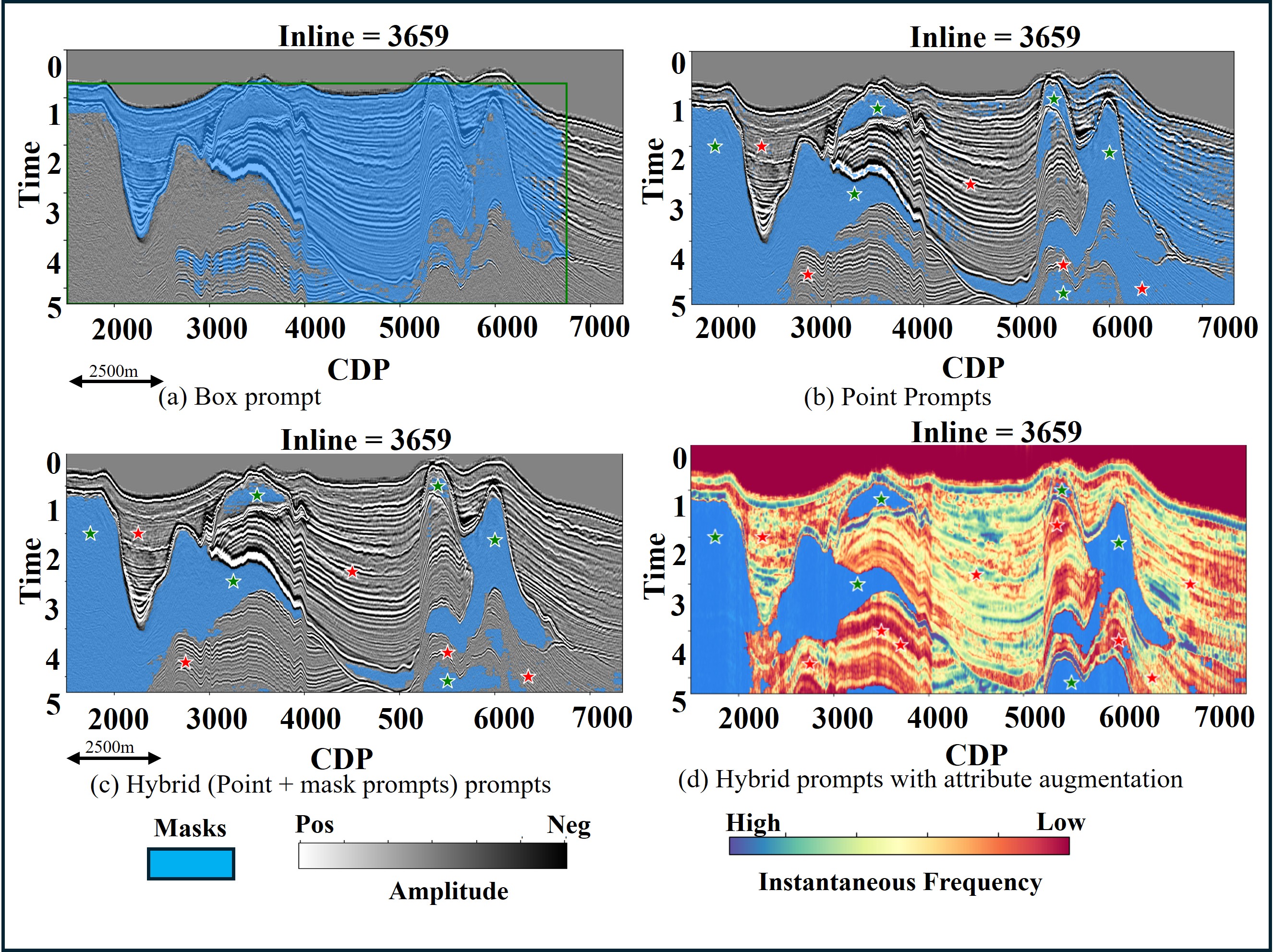}
    \caption{Comparison of prompting strategies and attribute-guided enhancement for salt segmentation. (a) Box prompt, (b) point prompts, (c) our hybrid point + mask prompting, and (d) attribute-enhanced segmentation using instantaneous frequency. Progressive improvement is observed from (a) to (c), with further refinement in (d) due to enhanced contrast at the salt–background interface.}
    \label{fig:fig_3}
\end{figure}

\begin{figure}
    \centering
    \includegraphics[width=0.7\linewidth]{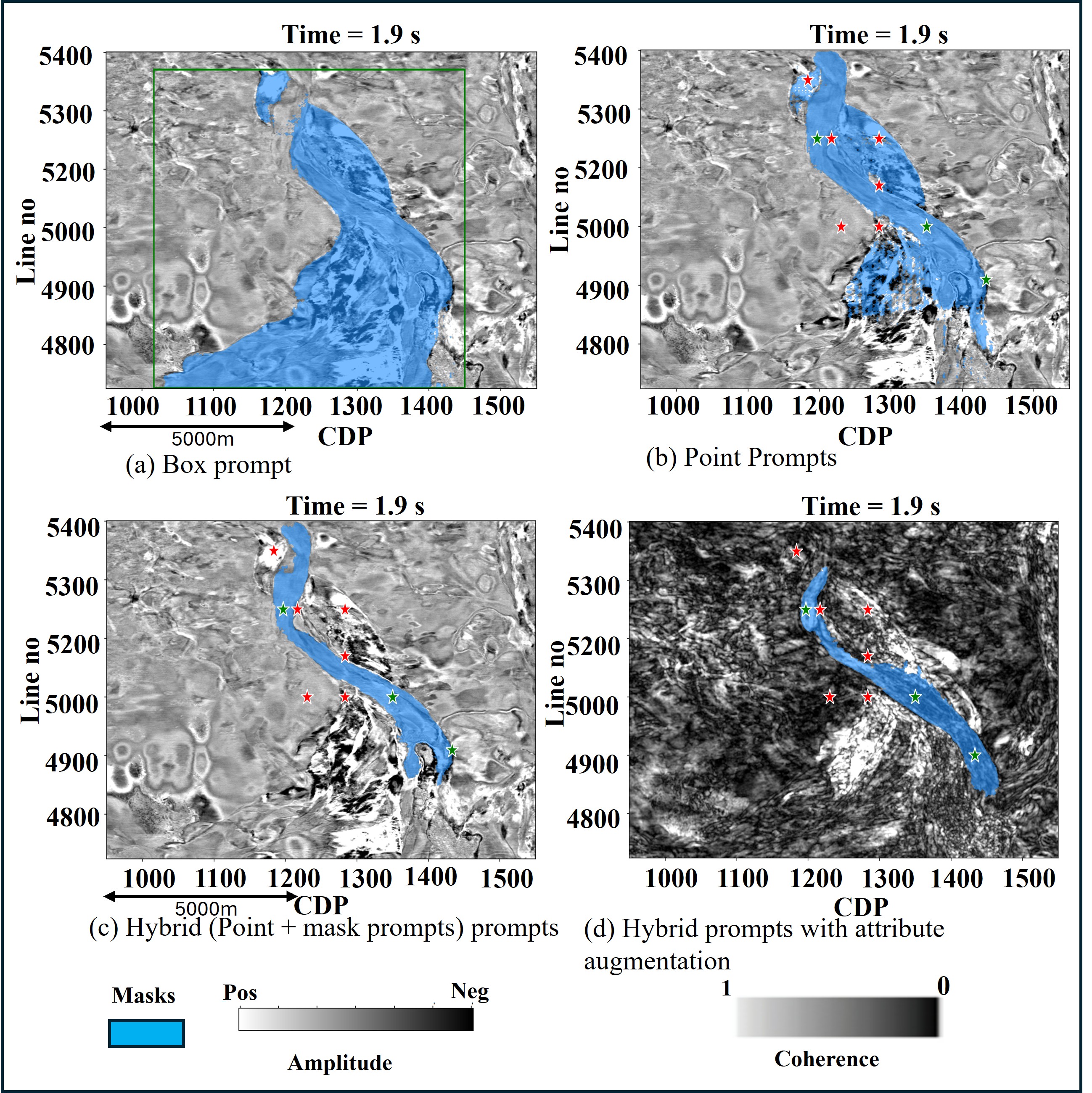}
    \caption{Comparison of prompting strategies and attribute-guided enhancement for channel segmentation. (a) Box prompt, (b) point prompts, (c) our hybrid point + mask prompting, and (d) attribute-enhanced segmentation using coherence. Improvement is observed from (a) to (c), with further refinement in (d) due to improved channel boundary visibility.}
    \label{fig:fig_4}
\end{figure}

The next set of experiments examined the extent to which seismic attributes can visually enhance specific geological features. We used instantaneous frequency for salt bodies, multi-spectral coherence and 50 Hz coherence for channels, and relative acoustic impedance for the Lower North Sea Group facies. As input to SAM, we replicated the relevant attribute across the three RGB channels. We then applied the same hybrid prompting strategy used in the previous experiments. A qualitative comparison of the results obtained with different prompting strategies and attribute-based enhancements is shown in Figures \ref{fig:fig_3}, \ref{fig:fig_4}, and \ref{fig:fig_5}.

Following traditional interpretation practices, we also conducted experiments to assess whether applying a colormap transformation to the original grayscale seismic sections could enhance the visual contrast of geological features and improve the model’s segmentation performance. Two popular colormaps were used for this purpose: the 'Seismic' and 'RdBu' colormaps from Python's Matplotlib library. The results are reported in Table \ref{tab:prompt-attribute-color} as well and shown in Figure \ref{fig:fig_6}. 

To further evaluate the effect of prompt density on segmentation performance, we analyzed precision, recall, and the F1-score for all three target classes. Performance is reported with 95 percent confidence intervals to ensure statistical reliability. Based on the ground-truth masks, we systematically generated foreground and background prompts. To prevent spatial clustering and ensure a representative distribution, we maintained a minimum distance of 100 pixels between any two prompts. Each experiment was repeated 20 times per dataset to account for stochastic variability in prompt placement. The mean values were then plotted against prompt count, as shown in Figure \ref{fig:fig_7}

\section{Discussion}

\subsection{Comparison of Prompting Strategies for Seismic Segmentation}

The effectiveness of different prompting strategies was evaluated on the salt and channel datasets, as summarized in Figures \ref{fig:fig_3} and \ref{fig:fig_4}, where each panel shows a progression from box prompts to point prompts, iterative point+mask prompting, and finally attribute-enhanced inputs. Box prompts (Figures \ref{fig:fig_3}a and \ref{fig:fig_4}a), although simple to implement, are poorly suited for seismic interpretation due to the irregular geometries of geological features. Because these features rarely conform to rectangular shapes, the box prompt often forces the model to include surrounding strata, resulting in significant over-segmentation. This behavior is particularly evident for salt bodies, where segmentation quality remains very low with an F1-score of 0.22. A similar limitation is observed for channels, where a single bounding box inevitably covers multiple branches of the fluvial system, leading to mixed segmentation and an F1-score of 0.62. In contrast, for Lower North Sea Group facies, box prompts perform relatively better for laterally continuous facies, where the target geometry conforms more closely to the rectangular prompt, yielding an F1-score of 0.77.

Point-based prompting, as shown in Figure \ref{fig:fig_3}b and \ref{fig:fig_4}b, improves segmentation by allowing interpreter-guided inclusion and exclusion through foreground and background points. This approach provided better localization of geological features and yielded strong performance for facies, increasing the F1-score to 0.90. However, for geometrically complex targets such as salt bodies and channel systems, point prompts alone remain insufficient, with F1-scores of 0.69 and 0.66, respectively, due to boundary leakage and incomplete delineation. To address this, we employed an iterative prompting strategy in which the model’s confidence logits are reused as mask prompts alongside the original point inputs. This sparse–dense prompting framework provided additional spatial guidance and substantially improved segmentation quality as shown in Figures \ref{fig:fig_3}c and \ref{fig:fig_4}c, increasing the F1-score from 0.69 to 0.90 for salt and from 0.66 to 0.78 for mud-filled channels. The Lower North Sea Group facies shows only a marginal improvement in segmentation performance, with the F1-score increasing from 0.90 to 0.91. We therefore do not present its results in a separate Figure, as segmentation remains consistently strong across all prompting strategies and exhibits only limited qualitative variation. These results are summarized quantitatively in Table~\ref{tab:prompt-attribute-color}. The final panels (Figures~\ref{fig:fig_3}d and \ref{fig:fig_4}d) further demonstrate that incorporating attribute-enhanced inputs within the same prompting framework provides an additional level of refinement for complex targets, as discussed in the following subsection.

\subsection{Attribute-Guided Enhancement of Prompting and Effect of Color Rendering}

We evaluated the effect of incorporating seismic attributes within the iterative prompting framework by comparing segmentation results obtained from raw seismic amplitude and attribute-guided inputs under identical prompting conditions.

For isolated salt bodies, instantaneous frequency provides a more informative representation of complex structural boundaries. Because salt flanks often appear as diffuse or low-coherence zones in amplitude data, the frequency-based representation enhances the contrast at salt–background interfaces, allowing the model to resolve irregular geometries better. When combined with the iterative prompting framework, this leads to a clear improvement in segmentation performance, with the F1-score increasing from 0.69 to 0.91, as shown in Figure~\ref{fig:fig_3}d. This result demonstrates that seismic attribute-enhanced inputs can substantially improve the delineation of low-contrast and structurally complex features that are poorly resolved in raw seismic amplitude.

\begin{figure}[htbp]
    \centering
    \includegraphics[width=0.9\linewidth]{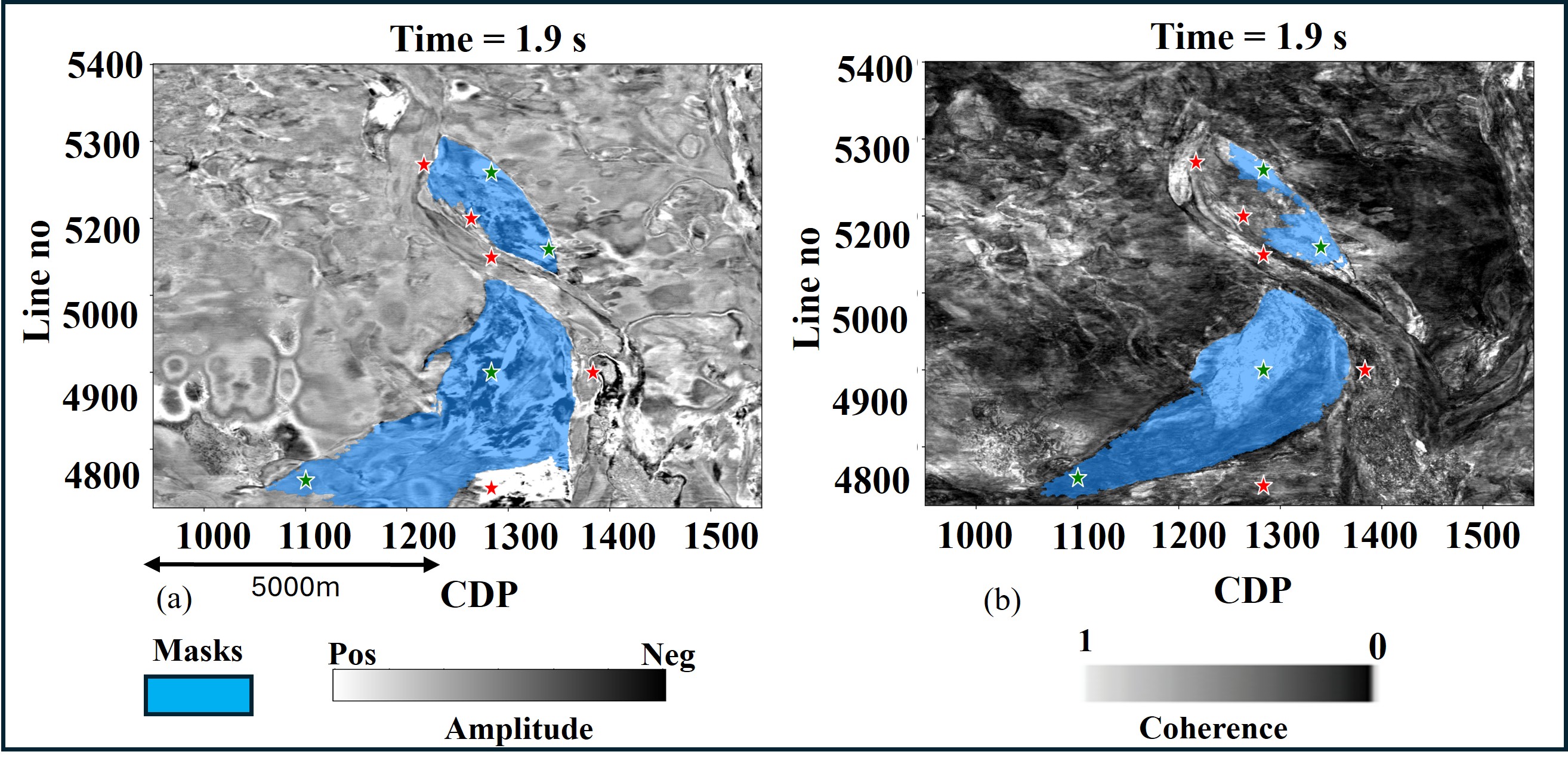}
    \caption{Comparison of sand-filled channel segmentation on (a) raw seismic amplitude and (b) multispectral coherence. The use of multispectral coherence significantly improved internal architecture resolution, increasing the F1-score from 0.66 to 0.84.}
    \label{fig:fig_5}
\end{figure}

For channel systems, coherence-based attributes enhance different aspects of fluvial morphology depending on their spectral content. High-frequency coherence (Figure~\ref{fig:fig_4}d) is applied to a narrow mud-filled channel, where it improves the delineation of fine-scale boundaries that appear blurred in the broadband seismic amplitude. This resulted in a clear improvement in segmentation quality, with the F1-score increasing from 0.62 to 0.84. To further demonstrate that this behavior is not limited to a single example, we examine an additional high-amplitude, sand-filled channel with a broader geometry using multispectral coherence (Figure \ref{fig:fig_5}). In this case, the attribute highlights structural continuity and channel architecture more effectively, improving the F1-score from 0.66 to 0.84. Together, these results show that different coherence representations selectively enhance geological features at different scales, enabling the iterative prompting framework to capture both narrow stratigraphic elements and broader channel architectures.

\begin{figure}[htbp]
    \centering
    \includegraphics[width=0.9\linewidth]{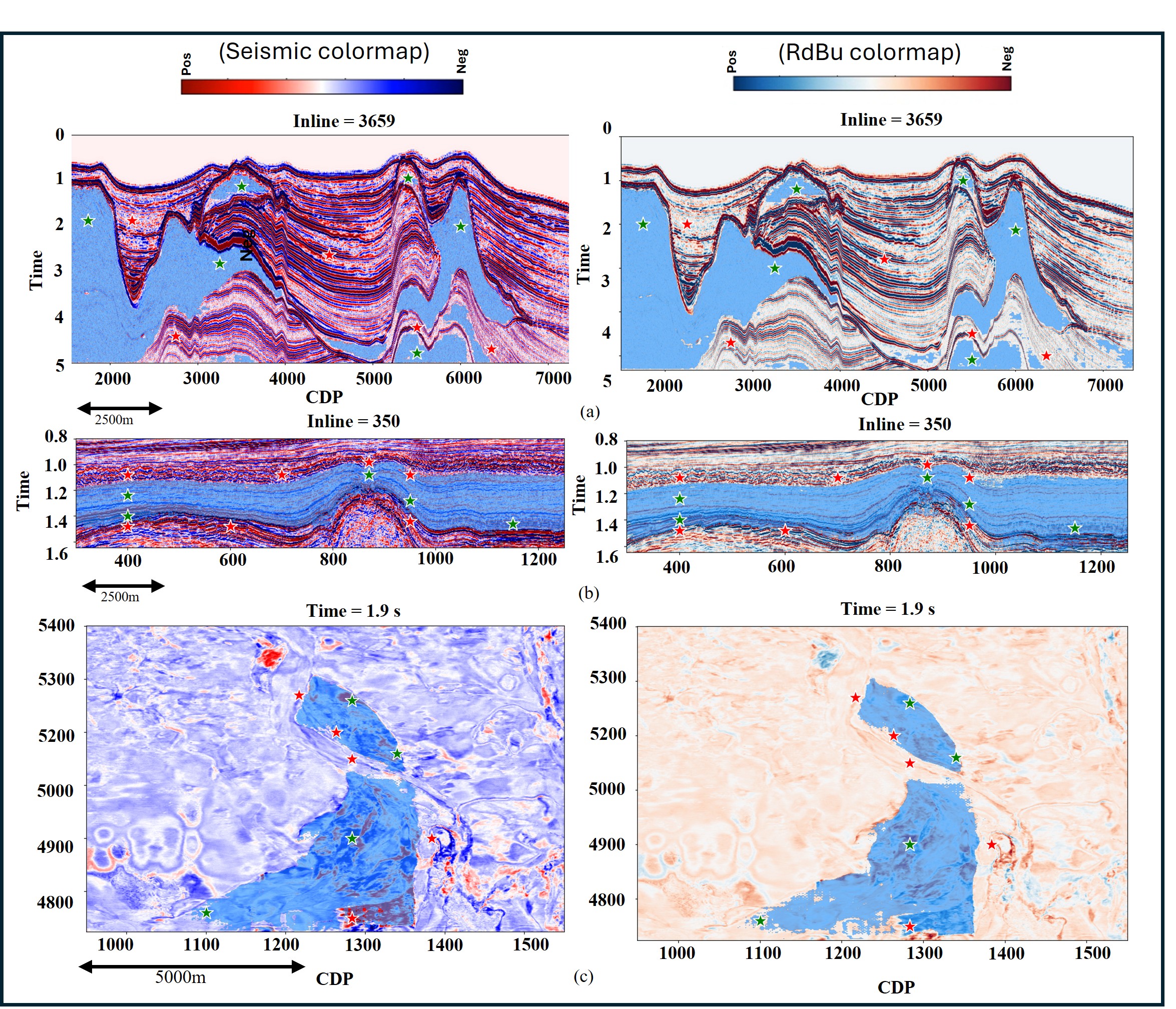}
    \caption{SAM segmentation results under two color renderings (Seismic and RdBu) for each test case. For salt bodies (a), colormap rendering improved segmentation performance by enhancing the strong amplitude contrast and sharp boundary definition of the target. In contrast, for facies (b) and channels (c), colormap rendering reduced segmentation performance by introducing visually prominent but geologically irrelevant gradients that obscured the spatial continuity required to delineate stratigraphic boundaries.}
    \label{fig:fig_6}
\end{figure}

For salt bodies, both the standard "Seismic" and "RdBu" colormaps yielded strong results (Figure \ref{fig:fig_6}a), achieving F1-scores of 0.920 and 0.927, respectively. These scores are comparable to those obtained using seismic attribute-enhanced representations, suggesting that colormap rendering can be effective when the geological target is structurally defined by strong amplitude contrast and sharp boundaries. For salt bodies, the interface is already clearly expressed in the seismic response, and color rendering further amplifies this physically meaningful contrast, enabling SAM to identify the target boundary more effectively. In contrast, for the Lower North Sea Group facies and the mud-filled channel (Figure \ref{fig:fig_6}b--c), the transition from grayscale amplitude to color reduced segmentation performance. Unlike salt, these stratigraphic targets are not primarily defined by strong amplitude magnitude but by subtle spatial continuity, internal texture, and lateral geological coherence. As a result, color remapping does not enhance the true defining signal; instead, it can exaggerate local amplitude fluctuations, thereby introducing visually salient but geologically irrelevant gradients and causing the model encoder to interpret spurious contrasts as boundaries. Accordingly, the F1-score for the Lower North Sea Group facies decreased to 0.87 with the Seismic colormap and 0.81 with RdBu, while the mud-filled channel F1-score decreased to 0.74 and 0.73, respectively. These results lead to an important practical guideline that colormap rendering is appropriate for structurally defined, high-contrast targets such as salt bodies. They should be used cautiously for stratigraphic interpretation where boundary definition depends on spatial continuity rather than amplitude magnitude alone.

\subsection{Impact of Prompt Density on Model Performance}

\begin{figure}
    \centering
    \includegraphics[width=0.8\linewidth]{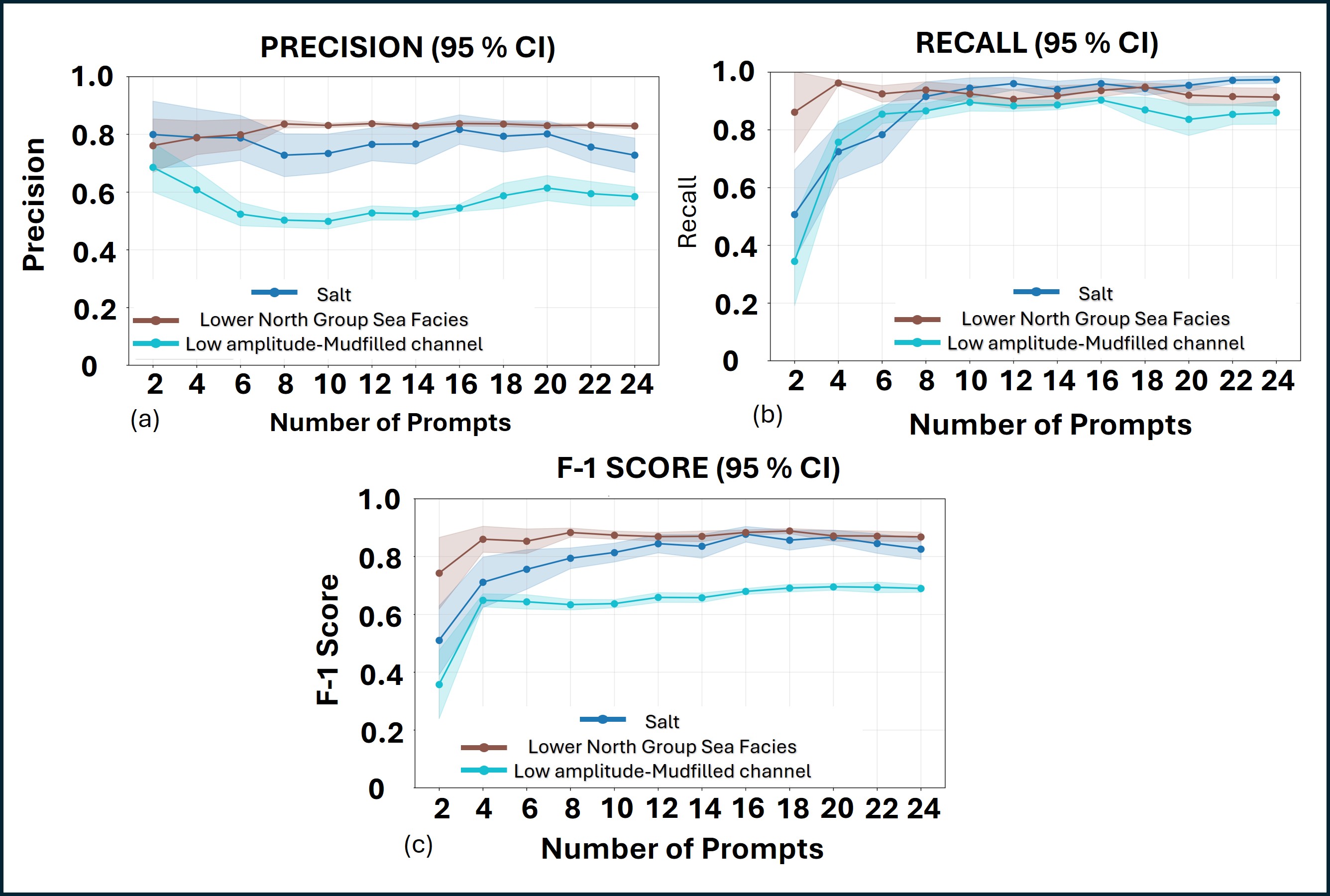}
    \caption{Quantitative evaluation of prompt density across salt, facies, and channel datasets. Increasing the number of prompts leads to rapid improvements in recall and stable F1-scores, with performance for most classes reaching a plateau at 8 to 12 prompts. While salt and facies achieve F1 scores exceeding 0.85, channel segmentation maintains lower precision due to boundary complexity, reaching a saturation point near 0.70. Confidence intervals (95 percent) represent the mean performance across 20 stochastic iterations.}
    \label{fig:fig_7}
\end{figure}

For channels and facies, precision, recall, and F1-score demonstrated a consistent prompt-conditioning effect across all classes. For salt and facies, precision remains high ($0.78$–$0.85$) as prompt counts increase. Meanwhile, recall improves rapidly with additional guidance, resulting in stable F1-scores exceeding $0.85$ that plateau at approximately 8–12 prompts. Lower North Sea Group facies achieved the highest overall F1 performance ($0.87$–$0.89$), followed closely by salt ($0.85$–$0.88$) as shown in Figure \ref{fig:fig_7}.

In contrast, channel segmentation exhibits lower precision ($0.50$–$0.60$), despite a substantial increase in recall (from $0.35$ to $0.85$) as the prompt count increases. This leads to a moderate F1 plateau near $0.70$. Due to random initialization of prompts each time, the model struggled with mud-filled channel segmentation; however, using seismic attributes and proper geoscientist-guided prompting, the model achieved the F1 score of $0.84$ (Table \ref{tab:prompt-attribute-color})

Across all classes, performance gains diminished markedly beyond approximately 12 random prompts, indicating a saturation point at which increasing prompt density no longer leads to meaningful improvements in segmentation accuracy.

\begin{table}[htbp]
\centering
\renewcommand{\arraystretch}{1.2}
\setlength{\tabcolsep}{4pt}

\resizebox{\linewidth}{!}{%
\begin{tabular}{|p{3cm}|
|c|c|c
|c|c|c
|c|c|c
|c|c|c
|c|c|}
\hline

\textbf{Feature}
& \multicolumn{3}{c|}{\textbf{Box}}
& \multicolumn{3}{c|}{\textbf{Point}}
& \multicolumn{3}{c|}{\textbf{Point+Mask}}
& \multicolumn{3}{c|}{\textbf{Attribute}}
& \multicolumn{2}{c|}{\textbf{Color}} \\

\cline{2-15}

& P & R & F1
& P & R & F1
& P & R & F1
& P & R & F1
& Seis & RdBu \\

\hline

Salt
& 0.14 & 0.43 & 0.22
& 0.54 & 0.96 & 0.69
& 0.89 & 0.92 & \textbf{0.90}
& 0.95 & 0.88 & \textbf{0.91}
& \textbf{0.920} & \textbf{0.927} \\
\hline

Mud-filled channel
& 0.47 & 0.92 & 0.62
& 0.58 & 0.76 & 0.66
& 0.74 & 0.82 & \textbf{0.78}
& 0.86 & 0.83 & \textbf{0.84}
& 0.747 & 0.737 \\
\hline

Lower North Sea Group Facies
& 0.66 & 0.94 & 0.77
& 0.86 & 0.96 & 0.90
& 0.87 & 0.96 & \textbf{0.91}
& 0.82 & 0.97 & 0.89
& 0.874 & 0.812 \\
\hline

\end{tabular}%
}

\caption{Quantitative comparison of segmentation performance across prompting strategies, attribute-guided inputs, and color rendering. Values are reported as Precision (P), Recall (R), and F1-score. Attribute results correspond to instantaneous frequency for salt, coherence at 50Hz for mud-filled channels, and relative acoustic impedance for Lower North Group Sea facies.}
\label{tab:prompt-attribute-color}
\end{table}

\subsection{Limitations of SAM}

Despite its strong performance, SAM possesses inherent limitations for seismic interpretation, most notably its lack of inherent geological context. Unlike a human interpreter, the model does not understand depositional principles or structural principles; it operates purely on image-based contrast and texture. Consequently, segmentation quality is highly sensitive to prompting strategy, where suboptimal placement can lead to significant mask leakage into adjacent facies or fragmented results. This dependency means the model's reliability is strictly capped by the user's ability to provide geologically meaningful prompts, particularly in areas where seismic boundaries are obscured by noise or acquisition artifacts.

Furthermore, the model’s performance degrades in visually ambiguous environments, such as low-contrast thin beds or zones with weak reflectivity. In these scenarios, the saturation effect observed in our prompt density analysis suggests that simply increasing the number of prompts does not always resolve boundary uncertainty.

\subsection{Conclusion}

In this study, we demonstrated that the Segment Anything Model (SAM) can be successfully applied to segment diverse geological features within seismic datasets. By leveraging SAM’s zero-shot generalization capability, we demonstrated that a foundation model originally trained on natural images can be effectively applied to complex geophysical interpretation without retraining. More importantly, this study established a principled zero-shot adaptation framework in which segmentation performance depends on combining geologic-target-aware seismic representations with hybrid prompting, rather than on task-specific model fine-tuning alone. Our findings suggest that segmentation success is influenced less by changes to the model itself and more by the interpreter’s skill in guiding the model through interactive prompting. High-contrast features such as salt bodies were often resolvable with minimal interaction and appropriate colormap selection. Complex stratigraphic features such as channel systems may require the synergy of iterative point-plus-mask prompting and attribute-guided inputs to get stable, geologically meaningful results. Across all tested geological targets, the proposed framework achieved F1-scores of up to 0.91 for salt bodies, 0.84 for channel systems, and 0.91 for facies, demonstrating competitive performance in a fully zero-shot setting without any domain-specific retraining.  

Ultimately, this workflow reframed the challenge of seismic interpretation from building separate deep learning models for each target class to developing a reusable interactive framework that incorporates geological expertise at runtime. By replacing the traditional bottleneck of training and maintaining geologic-feature-specific models with a single adaptable zero-shot workflow, we offer a more efficient and scalable approach to seismic interpretation. However, our results also emphasize that SAM is most effective when viewed as an assistant to the interpreter rather than as a fully autonomous solution. It is an interactive tool that can accelerate repetitive tasks and improve segmentation efficiency, but it still requires rigorous geoscientific validation. This interactive interpretation paradigm ensures that the efficiency of foundation models is balanced with the interpretive judgment required for robust geological characterization. Future work should explore extending this framework to full 3D volumetric segmentation, where propagating interpreter prompts across adjacent slices could further reduce annotation effort while maintaining geological consistency.

\section{ACKNOWLEDGMENTS}

We thank the Society of Exploration Geophysicists (SEG), New Zealand Petroleum and Minerals (NZPM), and dGB Earth Sciences for providing the seismic datasets. We also thank the AASPI lab and the AASPI consortium sponsors for their support. We extend our heartfelt appreciation to Dr. Kurt Marfurt and Dr. David Lubo-Robles for their insightful discussions and valuable suggestions.


\newpage

\bibliographystyle{seg}  
\bibliography{example}

\end{document}